\documentclass[letterpaper, 10 pt, conference]{ieeeconf}  

\IEEEoverridecommandlockouts                              

\usepackage{graphics} 
\usepackage{epsfig} 
\usepackage{times} 
\usepackage{amsmath} 
\usepackage{amssymb}  
\usepackage{afterpage}

\usepackage{graphicx}
\usepackage{upgreek}
\usepackage{bbm}
\usepackage{dsfont}
\usepackage{caption}
\usepackage{subcaption}
\usepackage{color}
\usepackage{hyperref}

\usepackage{booktabs}
\usepackage{makecell}
\usepackage{titlesec}
\usepackage{wrapfig}
\usepackage{array}
\usepackage{multirow}
\usepackage{colortbl}
\usepackage{tabularx}
\usepackage{cite}
\usepackage{mathabx}

\usepackage{xcolor}
\usepackage{pifont}

\title{\LARGE \bf
Track Any Motions under Any Disturbances
}

\author{Zhikai Zhang$^{*1,3}$ \quad Jun Guo$^{*1,3}$ \quad Chao Chen$^{2,3}$ \quad Jilong Wang$^{2,3}$ \quad Chenghuai Lin$^{3}$ \quad Yunrui Lian$^{1,3}$ \\ Han Xue$^{1,3}$ \quad  Zhenrong Wang$^{3}$ \quad Maoqi Liu$^{3}$ \quad Jiangran Lyu$^{2,3}$ \quad Huaping Liu$^{1}$ \quad He Wang$^{2,3}$ \quad Li Yi$^{\dag 1,4}$ 
\thanks{*Equal Contributions, $\dag$Corresponding Author}
\thanks{$^{1}$Tsinghua University, $^{2}$Peking University, $^{3}$Galbot, $^{4}$Shanghai Qi Zhi Institute}
\thanks{Paper website: \href{https://zzk273.github.io/Any2Track/}{https://zzk273.github.io/Any2Track/}}
}

\begin{document}

\makeatletter
\let\@oldmaketitle\@maketitle
\renewcommand{\@maketitle}{\@oldmaketitle 
  \centering
  \includegraphics[width=1\linewidth]{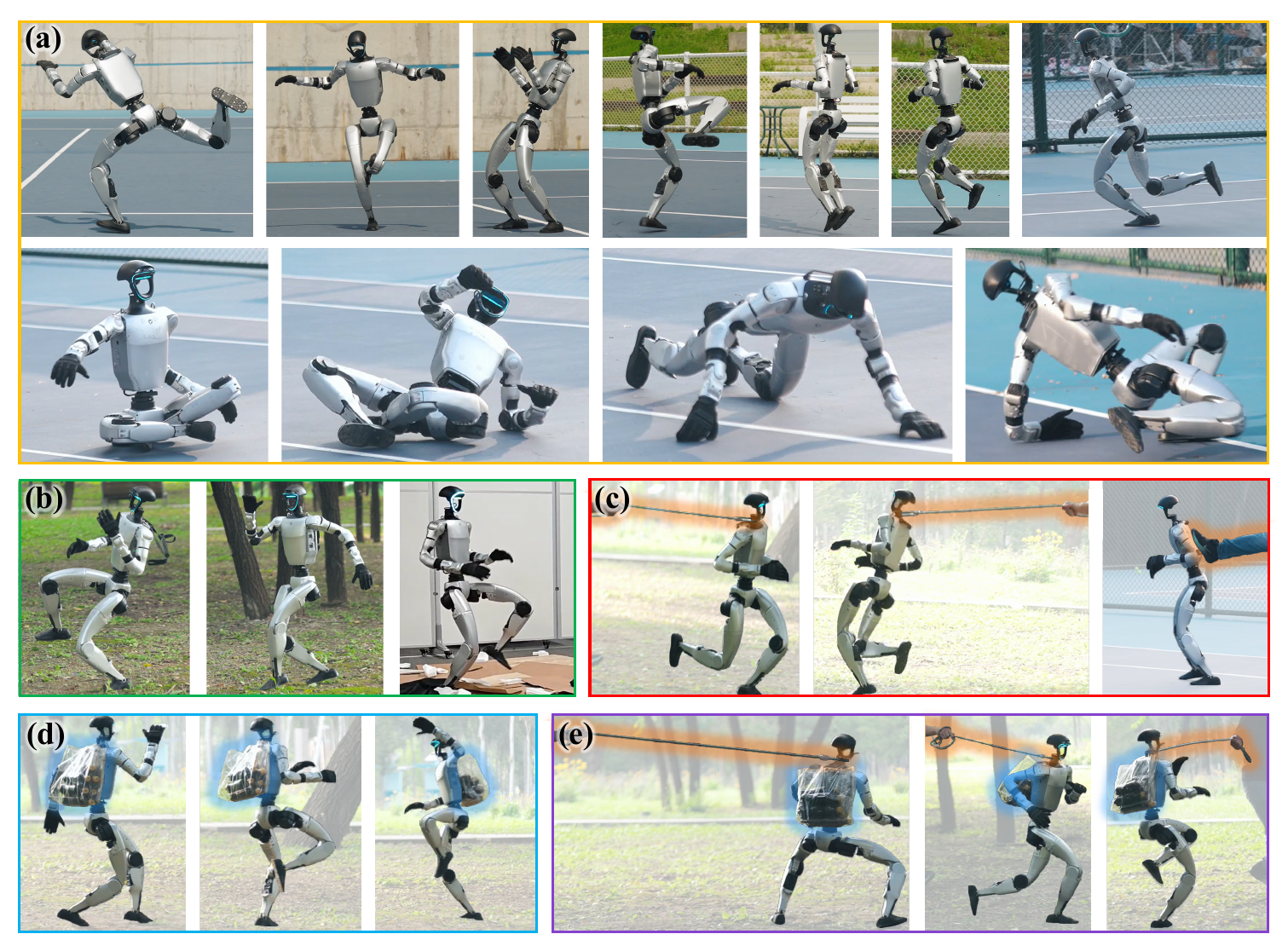}
  \captionof{figure}{\textbf{(a)} The humanoid tracks \textit{diverse}, \textit{highly dynamic}, and \textit{contact-rich} motions using a single policy. \textbf{(b)} The humanoid tracks motions on different \textit{terrains}. \textbf{(c)} The humanoid tracks motions against different \textit{external forces}, including pulling forces from the rope and pushing forces from the human feet. \textbf{(d)} The humanoid tracks motions against \textit{physical property changes} (the payload on the back).
  \textbf{(e)} The humanoid tracks motions against \textit{all dynamics disturbances}, including \textit{terrains}, \textit{external forces}, and \textit{physical property changes} simultaneously.  
  }
  \label{fig:teaser}
  \vspace{-0.4cm}
}
\makeatother

\maketitle
\thispagestyle{empty}
\pagestyle{empty}

\setcounter{figure}{1}

\begin{abstract}

A foundational humanoid motion tracker is expected to be able to track diverse, highly dynamic, and contact-rich motions. More importantly, it needs to operate stably in real-world scenarios against various dynamics disturbances, including terrains, external forces, and physical property changes for general practical use. To achieve this goal, we propose Any2Track (Track Any motions under Any disturbances), a two-stage RL framework to track various motions under multiple disturbances in the real world. Any2Track reformulates dynamics adaptability as an additional capability on top of basic action execution and consists of two key components: AnyTracker and AnyAdapter. AnyTracker is a general motion tracker with a series of careful designs to track various motions within a single policy. AnyAdapter is a history-informed adaptation module that endows the tracker with online dynamics adaptability to overcome the sim2real gap and multiple real-world disturbances. We deploy Any2Track on Unitree G1 hardware and achieve a successful sim2real transfer in a zero-shot manner. Any2Track performs exceptionally well in tracking various motions under multiple real-world disturbances, as shown in Figure~\ref{fig:teaser}. For real-world demos, please refer to \href{https://zzk273.github.io/Any2Track/}{https://zzk273.github.io/Any2Track/}.

\end{abstract}
\section{INTRODUCTION}
Humanoid motion tracking~\cite{fu2024humanplus,he2024omnih2o,ji2024exbody2,ze2025twist,chen2025gmt,he2025asapaligningsimulationrealworld} aims to reproduce human motion sequences on a humanoid platform, enabling more expressive and anthropomorphic motion behavior compared with pure RL-based controllers~\cite{ben2025homie,zhang2025unleashing,gu2024advancing,xue2025unified,sferrazza2024humanoidbench,radosavovic2024real,li2025hold,zhang2025falcon}. A foundational humanoid motion tracker should be able to track diverse, highly dynamic, and contact-rich motions to capture general human motion knowledge. In addition, it needs to operate stably in real-world scenarios against various disturbances, including terrains, external forces, and physical property changes (e.g., payload, friction) for general practical use. However, existing humanoid motion trackers fail to have general motion tracking capability and dynamics adaptability simultaneously. They exhibit varying levels of limitations in motion tracking capabilities and resilience to different real-world disturbances. The comparison of existing works and our method is shown in Table~\ref{table:overall_comparison}.

The key challenges to construct a fundamental humanoid motion tracker can be summarized in two points: 1) how to construct a unified general motion tracker to efficiently and with high quality learn diverse motion control strategies; 2) how to endow the controller with online dynamics adaptability to overcome the sim2real dynamics gap and various real-world disturbances. 

Towards the goal of constructing a foundational humanoid motion tracker, we propose a two-stage RL framework called \textbf{Any2Track} (\textbf{Track Any} motions under \textbf{Any} disturbances) in this work. Any2Track reformulates dynamics adaptability as an additional capability on top of basic action execution and consists of two key components: AnyTracker and AnyAdapter.

In the first stage, we construct \textbf{AnyTracker}, a general motion tracker 
to track diverse, highly dynamic, and contact-rich motions. We found that the bottleneck in training a general motion tracker lies in the complex action spaces brought by high degrees of freedom and motion diversity. Thus, we propose a series of careful designs, including canonicalized action spaces and a specialist-to-generalist strategy, to alleviate the optimization difficulties brought by the complex action spaces. AnyTracker is trained as a base policy without any dynamics randomization to avoid tracking performance degradation.

In the second stage, we introduce dynamics variance and further propose a history-informed adaptation module on top of AnyTracker, which is called \textbf{AnyAdapter}. AnyAdapter endows the base tracker with online dynamics adaptability for different kinds of disturbances, including terrains, external forces, and physical property changes, without compromising its foundational expressive motion tracking capability. AnyAdapter utilizes dynamics-aware world model prediction as a proxy task to extract dynamics features as neural embeddings from the history buffer, which can serve as an informative representation to identify environment dynamics. Rather than directly finetuning the original parameters of the base tracker for dynamics adaptability, which may harm the tracking skills already acquired, we freeze the base tracker and introduce an adapter~\cite{hu2022lora} architecture.
The adapter is trained to adaptively adjust the base policy's actions to accommodate different environments, taking dynamics embeddings as input.

We deploy Any2Track on Unitree G1 hardware and achieve a successful sim2real transfer in a zero-shot manner. Our method shows impressive results in tracking diverse, highly dynamic, and contact-rich motions when faced with real-world disturbances from different sources, including terrains, external forces, and physical property changes. Extensive experiments are conducted both in the simulation and in the real world to validate the effectiveness of our major designs. 

Our contributions are fourfold:

\begin{itemize}
\item \textbf{Any2Track}: A novel two-stage RL framework for robust humanoid motion tracking that excels under significant real-world disturbances.
\item \textbf{AnyTracker}: A general motion tracker with a series of careful designs to alleviate the optimization difficulties brought by complex action spaces and track diverse, highly dynamic, and contact-rich motions.
\item \textbf{AnyAdapter}: A history-informed module that provides online adaptation to dynamic shifts and external perturbations, including terrains, external forces, and physical property changes.
\item \textbf{State-of-the-art Performance}: We demonstrate superior real-world tracking under disturbances on the Unitree G1 hardware, validated by extensive experiments in both simulation and reality.

\end{itemize}

\begin{table*}
\centering
    \renewcommand{\arraystretch}{1.2}
    \vspace{0.2cm}
    \caption{\textbf{Overall comparison of different motion tracking methods.} We compare Any2Track with existing methods according to their performance in publicly available demos and their proposed methods. Any2Track demonstrates unprecedented levels of motion diversity, high dynamism, and contact complexity. More distinctively, our method adapts to multiple real-world disturbances for general practical use. It is worth noting that, although both Any2Track and GMT~\cite{chen2025gmt} use the combination of LAFAN1~\cite{harvey2020robust} and AMASS~\cite{mahmood2019amass} as the training dataset, GMT removed a significant portion of highly dynamic or contact-rich motions. In contrast, we retained these motions.}
    \label{table:overall_comparison}
    \begin{tabular}
    {@{\extracolsep{\fill}} p{0.10\linewidth} 
    >{\centering\arraybackslash}p{0.16\linewidth} 
    >{\centering\arraybackslash}p{0.12\linewidth} 
    >{\centering\arraybackslash}p{0.08\linewidth}
    >{\centering\arraybackslash}p{0.0001\linewidth}
    >{\centering\arraybackslash}p{0.07\linewidth} 
    >{\centering\arraybackslash}p{0.1\linewidth} 
    >{\centering\arraybackslash}p{0.19\linewidth} 
    }  
    \hline
    \multirow{2}{*}{Method} & \multicolumn{3}{c}{Track Any Motions} & & \multicolumn{3}{c}{Adapt to Any Disturbances}\\
    \cline{2-4}
    \cline{6-8}
    & Diversity & Highly Dynamic & Contact-rich & & Terrain & External Force & Physical Property Change \\
    \hline
    Exbody2~\cite{ji2024exbody2}  & CMU dataset~\cite{cmu_mocap} & \textcolor{red}{\ding{55}} & \textcolor{red}{\ding{55}} & & \textcolor{red}{\ding{55}} & \textcolor{red}{\ding{55}} & \textcolor{red}{\ding{55}}\\
    ASAP~\cite{he2025asapaligningsimulationrealworld}  & One motion clip & \textcolor{green}{\ding{51}} & \textcolor{red}{\ding{55}}& & \textcolor{red}{\ding{55}} & \textcolor{red}{\ding{55}} & \textcolor{red}{\ding{55}}\\
    GMT~\cite{chen2025gmt}  & LAFAN1 + AMASS*& \textcolor{green}{\ding{51}}& \textcolor{red}{\ding{55}}& & \textcolor{red}{\ding{55}} & \textcolor{red}{\ding{55}} & \textcolor{red}{\ding{55}}\\
    \textbf{Ours} & \textbf{LAFAN1 + AMASS} & \textcolor{green}{\ding{51}} &\textcolor{green}{\ding{51}} & & \textcolor{green}{\ding{51}} & \textcolor{green}{\ding{51}} & \textcolor{green}{\ding{51}} \\
    \hline
    \end{tabular}
    
\vspace{-0.5cm}
\end{table*}
\section{RELATED WORKS}

\subsection{Motion Tracking}

Motion tracking aims to reproduce human motions to learn anthropomorphic whole-body controllers, which has been extensively studied in character animation~\cite{luo2023perpetual,peng2018deepmimic,zhang2024freemotion,tessler2024maskedmimic,luo2023universal,luo2024grasping,li2025learning,liu2024mimicking}. ~\cite{luo2023perpetual} shows progressive results in constructing a general motion tracking strategy with dynamic, versatile, and perpetual tracking capability. 

Recently, robotics researchers have started to explore the reproduction of human motions on real-world humanoid hardware platforms~\cite{fu2024humanplus,he2024omnih2o,ji2024exbody2,chen2025gmt,ze2025twist,he2025asapaligningsimulationrealworld}, to enable more expressive motion behavior compared with pure RL-based controllers~\cite{ben2025homie,zhang2025unleashing,gu2024advancing,xue2025unified}. However, compared with character animation, motion tracking on humanoid robots still lags behind significantly due to the cross-embodiment gap, actuator limits, and sim-to-real challenges. HumanPlus~\cite{fu2024humanplus} and OmniH2O~\cite{he2024omnih2o} mainly focus on quasi-static loco-manipulation tasks. ASAP~\cite{he2025asapaligningsimulationrealworld} achieves highly dynamic motion tracking. However, it overfits to motion clips of only a few seconds. GMT~\cite{chen2025gmt} constructs a general motion tracker with multiple careful designs in motion distribution sampling and network architecture. Although this method can track a considerable portion of data from LAFAN1~\cite{harvey2020robust} and AMASS~\cite{mahmood2019amass} datasets, it shows limited capability when dealing with highly dynamic and contact-rich motions. Moreover, GMT does not show adaptability to external disturbances, which limits its practicality in real-world scenarios.  

Compared with existing works, Any2Track demonstrates unprecedented levels of motion diversity, high dynamism, and contact complexity within a unified tracking policy. More distinctively, our method endows the motion tracker with online dynamics adaptability to overcome the sim-to-real gap and various real-world disturbances, marking a significant step toward general real-world use.

\subsection{Online Dynamics Adaptation for Legged Robots}
A key challenge in constructing a fundamental humanoid motion tracker is to overcome the sim2real dynamics gap and various real-world disturbances for general practical use. Existing works~\cite{chen2025gmt,fu2024humanplus,ji2024exbody2} often rely on na\"ive domain randomization in the hope that the policy can perform robustly under different environment dynamics. However, due to the unawareness of the environment dynamics, the policy cannot adaptively adjust its actions, causing it to behave conservatively when faced with large dynamic variance.

To cope with more complex real-world environments for legged robots, online dynamics adaptation methods~\cite{gu2024advancing,li2025reinforcement,kumar2021rma,lyu2025dywa} are needed. Existing works~\cite{gu2024advancing,li2025reinforcement,kumar2021rma} utilize robot-environment interaction history to adaptively estimate real-world dynamics and adjust robot behavior accordingly. Rapid Motor Adaptation (RMA)~\cite{kumar2021rma} 
learns to encode recent state histories into a latent representation of environment dynamics by distilling knowledge from a privileged teacher encoder.
Li \textit{et al.}~\cite{li2025reinforcement} designs a dual-history architecture to overcome the sim2real dynamics gap with history information over varying horizons. DWL~\cite{gu2024advancing} utilizes a denoising world model to predict environment configuration and privileged robot states simultaneously from history information.

Compared with existing works, we take a further step in online dynamics adaptation.
To learn better dynamics representations from the history buffer, we introduce dynamics-aware world model prediction as a proxy task. The learned representations are more informative in dynamics. 
To decouple the acquisition of motion tracking capability and dynamics adaptability to reduce the difficulty of joint optimization, we introduce an adapter architecture and a finetuning stage on top of a base tracker to accommodate different environment dynamics.

\section{TRACK ANY MOTIONS UNDER ANY DISTURBANCE}
\begin{figure*}
    \centering
    \vspace{0.1cm}
    \includegraphics[width=1.0\linewidth]{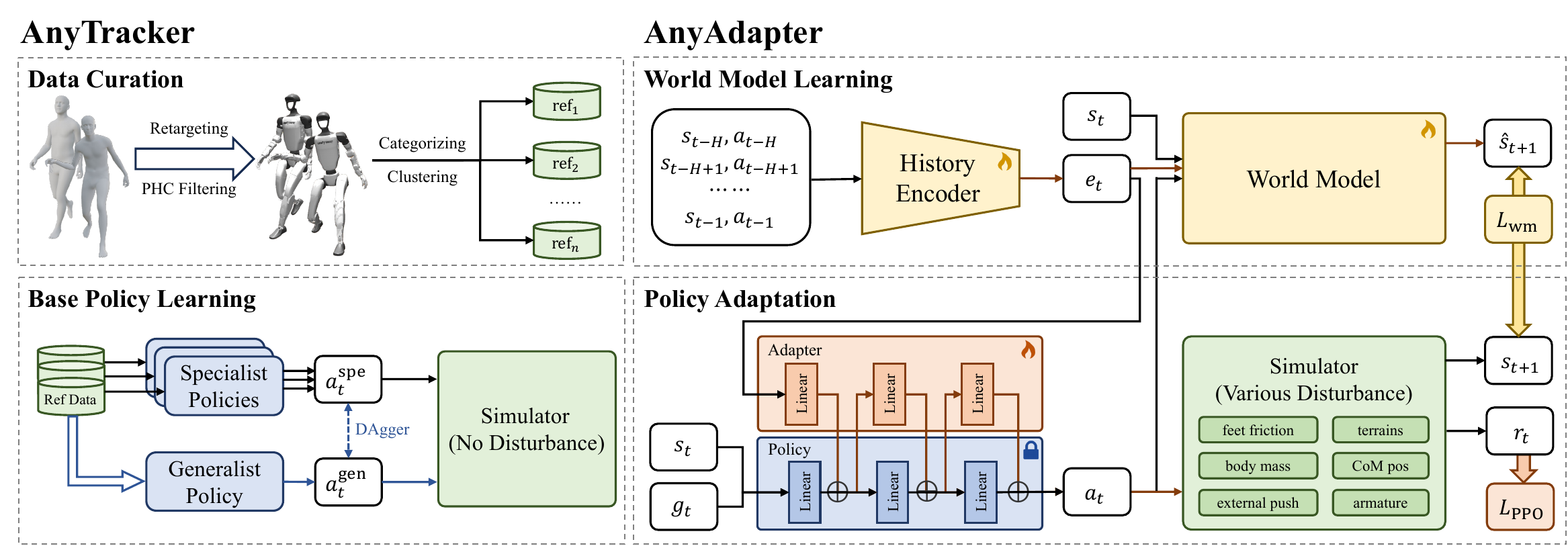}
    \caption{\textbf{Overview of our method.} \textbf{Any2Track} consists of two key components: AnyTracker and AnyAdapter. \textbf{AnyTracker} is a general motion tracker with a series of careful designs. \textbf{AnyAdapter} is a history-informed adaptation module on top of AnyTracker. AnyAdapter endows the base tracker with online dynamics adaptability without compromising its fundamental expressive motion tracking capability.}
    \label{fig:pipeline}
\vspace{-0.6cm}
\end{figure*}

We propose Any2Track in this work, a two-stage RL framework to build a foundational humanoid motion tracker. The design philosophy of Any2Track is to reformulate dynamics adaptability as an additional capability on top of basic action execution and decouple its learning process. Such a design can demonstrate exceptional robustness to environmental disturbances while preserving its motion expressiveness.
Concretely, we first construct AnyTracker, a general motion tracker with a series of careful designs to alleviate the optimization difficulties brought by complex action spaces and track various motions, as shown in Section~\ref{sec:anytracker}. At this stage, AnyTracker is trained as a base policy without any dynamics randomization to avoid tracking performance degradation. We then propose AnyAdapter, a history-informed adaptation module to overcome the sim2real dynamics gap and various real-world disturbances, as shown in Section~\ref{sec:anyadapter}. At this stage, we introduce dynamics domain randomization and fine-tune the motion tracker for dynamics adaptability. 
The pipeline is shown in Figure~\ref{fig:pipeline}. We use PPO~\cite{ppo} as our reinforcement learning framework and MuJoCo~\cite{todorov2012mujoco,zakka2025mujoco} for simulation. We train our policy in parallel on 8 GPUs.

\subsection{AnyTracker: Track Any Motions}
\label{sec:anytracker}

Our tracker can be seen as an RL policy \(\pi: \mathcal G \times \mathcal S \mapsto \mathcal A\), which maps humanoid proprioception state and tracking goals to low-level robot actions. At each timestep $t$, the inputs of policy $\pi$ consist of current state $s_{t}$ and current tracking goals $g_{t}$. $s_{t}$ includes angular velocity, projected gravity, per-joint position, per-joint velocity, and last-frame action. $g_{t}$ depicts next-frame humanoid target motion, including target per-joint position, target per-joint velocity, and target rigid body information in the local frame. The policy $\pi$ needs to output per-joint action $a_{t}$, which is further fed into a PD controller to compute actuator torques. The task goal of motion tracking is that at timestep ${t+1}$, the robot's state $s_{t+1}$ should be as close as possible to the target motion depicted by $g_{t}$ while maintaining balance and safety. The reward terms we used are shown in Table~\ref{table:rewards}.

To achieve general motion tracking, we use the combination of AMASS~\cite{mahmood2019amass} and LAFAN1~\cite{harvey2020robust} motion datasets as our training data.
We filter out motions infeasible for tracking following PHC~\cite{luo2023perpetual}, primarily those involving interactions with an unavailable scene (e.g., climbing stairs). However, unlike GMT~\cite{chen2025gmt}, we retain highly dynamic and contact-rich motions. We found that with proper reward and policy observation design, training a motion tracker on a few highly dynamic or contact-rich motion trajectories is no longer difficult. The bottleneck in training a general motion tracker lies in the complex action spaces brought by the humanoid's high degrees of freedom and the motion diversity. For different joints and different motion categories, the action distributions vary significantly, making it challenging to learn all distributions well in a single RL optimization process. Therefore, we propose novel designs to alleviate this difficulty in terms of degrees of freedom and motion diversity, respectively. 

\begin{table}[]
\vspace{0.2cm}
\centering
\caption{\textbf{Reward terms used in AnyTracker.}}
\setlength{\tabcolsep}{2pt}
\begin{tabular}{@{}cccc@{}}
\toprule
Term                     & Weight               & Term                    & Weight               \\ \midrule
\multicolumn{4}{c}{Task}                                                                  \\ \midrule
Upper-body position      & $1.0$                  & Lower-body position     & $0.5$                  \\
Torso roll/pitch         & $1.0$                  & Body rotation           & $0.5$                  \\
Body linear velocity     & $0.5$                  & Body angular velocity   & $0.5$                  \\
DoF position             & $0.75$                 & DoF velocity            & $0.5$                  \\
Root linear velocity     & $1.0$                  & Root angular velocity   & $1.0$                  \\
Root height              & $1.0$                  & Feet height             & $1.0$                  \\ \midrule
\multicolumn{4}{c}{Regularization}                                                               \\ \midrule
Action rate              & $-0.5$                 & DoF velocity rate       & $-1\times 10^{-6}$ \\
Torques                  & $-2\times 10^{-5}$   &                         &                      \\ \midrule
\multicolumn{4}{c}{Penalty}                                                                      \\ \midrule
DoF position limits      & $-10.0$                & DoF velocity limits     & $-5.0$                 \\
Self collision           & $-10.0$                & Termination             & $-200.0$               \\ \bottomrule
\end{tabular}
\label{table:rewards}
\vspace{-0.5cm}
\end{table}

\subsubsection{Canonicalized Action Spaces} 
The huge diversity of each joint's action distribution makes it difficult for the policy to learn a general and high-quality control strategy. Thus, we propose to design a canonicalized motion tracking action space so that the policy only needs to predict a compact multi-joint action distribution. Concretely, we first utilize the $tanh$ function to map the policy prediction $\pi(a_{t}|s_{t})$ to the range $[-1,1]$. However, the action range varies for different joints, and using a unified policy prediction range is not suitable for all joints and can severely hinder the tracker's performance. Therefore, we use empirically designed hyperparameters $\mathbf{\alpha} \in \mathbb{R}^{\text{num\_joints}}$ to rescale each joint's action scale. Instead of directly predicting the PD targets, we predict residual PD offsets relative to the reference motion. The final PD targets can be written as:
\begin{equation}
q_{d}=\Tilde{q}_{t+1} + \alpha\tanh(\pi(a_{t}|s_{t})), 
\end{equation}
where $\Tilde{q}_{t+1}$ is the target next-frame joint position.

\subsubsection{Motion Clustering and Specialist-to-Generalist} 
Different categories of motions exhibit different motion patterns and action distributions, posing challenges for general motion tracker training.
Therefore, we propose to cluster motions according to categories and train a specialist for each motion cluster. By doing so, each specialist only needs to handle a set of motions with similar action distributions, which greatly reduces the training difficulty and improves the final results.
For LAFAN1~\cite{harvey2020robust}, as the motion categories are already provided, we train a specialist policy for each motion category. For AMASS~\cite{mahmood2019amass}, HumanML3D~\cite{guo2022generating} provides motion categories labels (the ``VERB'' term in the annotation). AMASS contains a large number of motion categories, and thus training a specialist for each category would be inefficient. Therefore, we feed the motion category labels into CLIP~\cite{radford2021learning} to compute text feature embeddings. Then we cluster motions based on these embeddings using the K-means~\cite{mcqueen1967some} method. We split AMASS into 6 subsets and train corresponding specialists. 
At last, we adopt the specialist-to-generalist paradigm~\cite{zhang2025unleashing} and use DAgger~\cite{ross2011reduction} to distill all specialists into a generalist policy.

\subsection{AnyAdapter: Adapt to Any Disturbances}
\label{sec:anyadapter}
Based on the AnyTracker acquired previously, we introduce environment dynamics variance in Table~\ref{table:dynamics} at this stage and propose AnyAdapter, a history-informed adaptation module for different kinds of disturbances.

\begin{table}[]
\vspace{0.2cm}
\centering
\caption{\textbf{Dynamics variance used in Any2Track.} We add three types of randomization, including terrains, external forces, and physical property changes. }
\setlength{\tabcolsep}{32pt}
\begin{tabular}{@{}cc@{}}
\toprule
Term                           & Value                       \\ \midrule
\multicolumn{2}{c}{Terrains}                                \\ \midrule
Floor friction                 & $\mathcal{U}(0.3, 2.0)$    \\
Max terrain height             & $0.3$                      \\
Noise scale                    & $\mathcal{U}(10.0, 16.0)$  \\
Noise octaves                  & $\mathcal{U}(5.0, 8.0)$    \\
Noise persistence              & $\mathcal{U}(0.3, 0.5)$    \\
Noise lacunarity               & $\mathcal{U}(2.0, 4.0)$    \\ \midrule
\multicolumn{2}{c}{External Forces}                            \\ \midrule
Interval range                 & $\mathcal{U}(5.0, 10.0)$   \\
Velocity magnitude range       & $\mathcal{U}(0.1, 1.0)$    \\ \midrule
\multicolumn{2}{c}{Physical Property Changes}                      \\ \midrule
DoF friction scaling           & $\mathcal{U}(0.5, 2.0)$    \\
Armature scaling               & $\mathcal{U}(1.0, 1.05)$   \\
Torso CoM position change   & $\mathcal{U}(-0.15, 0.15)$ \\
Torso mass change              & $\mathcal{U}(-3.0, 6.0)$   \\
Default DoF position jittering & $\mathcal{U}(-0.05, 0.05)$ \\ \bottomrule
\end{tabular}
\label{table:dynamics}
\vspace{-0.5cm}
\end{table}

We argue that there are two key components for online dynamics adaptability: 1) a good dynamics encoder that estimates adequate environment dynamics features from robot-environment interaction history buffer; 2) an effective training strategy that helps the control policy to properly leverage these dynamics features to adaptively adjust its behavior without sacrificing motion expressiveness. Thus, AnyAdapter utilizes dynamics-aware world model predication as a proxy task to extract dynamics features into neural embeddings from the history buffer, which can serve as an informative representation to tell different dynamics. To facilitate the motion tracker’s learning of dynamics adaptability, AnyAdapter refrains from coupling the basic motion-tracking capability with dynamics adaptability in a single network and instead introduces an additional adapter architecture. 

\subsubsection{Learn Informative Dynamics Embeddings} 
Given the history of robot–environment interactions, the motion tracker is expected to leverage this information to estimate the current environment dynamics and adapt its behavior accordingly. However, we found that the interaction history often contains substantial irrelevant information and noise, making direct utilization challenging for model learning. To address this, it is necessary to introduce proxy tasks for representation learning, enabling the extraction of informative feature embeddings that capture rich dynamics knowledge.
In this work, we propose dynamics-aware world model prediction, which contains a history encoder $\phi$ and a world model $\omega$. At each timestep $t$, the history encoder extracts dynamics feature embeddings from history information:
\begin{equation}
    e_{t} = \phi(h_{t}),
\end{equation}
where $h_{t}=\{s_{t-H},a_{t-H},...,s_{t-1},a_{t-1}\}$ and $H=79$ is the history window size. The world model needs to predict the next-frame robot state autoregressively:
\begin{equation}
    \hat{s}_{t+i+1} = \omega(\hat{s}_{t+i}, {a}_{t+i}, e_{t+i}),
\end{equation}
where $i\in\{0,1,...,N-1\}$ and $N=20$ is the autoregressive prediction window size.
Since we introduce environment dynamics variance at this stage, different environment configurations can significantly affect the forward dynamics. The history encoder needs to provide sufficiently informative dynamics features to identify environments, so that the world model can correctly predict the next-frame robot state taking the features as input (\textbf{dynamics-aware} world model prediction).

To train the history encoder $\phi$ and world model $\omega$, we first sample a window of $H+1+N=100$ state-action pairs from the data buffer. For initialization, the history encoder $\phi$ takes the first $H$ data pairs as history input and computes initial dynamics feature embeddings $e_{t}$. Then the world model takes the $(H+1)$-th state $s_{t}$ as the initial state $\hat{s}_t$, and autoregressively predicts the next $N$ states. The loss function can be written as:
\begin{equation}
    \mathcal{L}_{\text{wm}} = \sum_{i=1}^{N}\|s_{t+i}-\hat{s}_{t+i}\|_{1}.
\end{equation}
$\mathcal{L}_{\text{wm}}$ updates both the history encoder $\phi$ and the world model $\omega$ simultaneously through backpropagation.

\subsubsection{Dynamics Adaptability Injection}
Existing works~\cite{li2025reinforcement,kumar2021rma} often use a single network to handle both basic action execution and dynamics adaptability. Such coupling will increase the difficulty of network learning. When the dynamics vary excessively, basic action execution capability can be severely affected and tends to become over-conservative, which is highly detrimental to motion tracking tasks that demand expressiveness and anthropomorphic quality.

In this work, we reformulate dynamics adaptability as an additional capability on top of basic action execution. To avoid compromising the already acquired motion tracking capability. We freeze the network parameters of Any2Track and introduce an adapter~\cite{hu2022lora} architecture for fine-tuning. The adapter $\xi$ consists of $M$ layers with zero-initialized weights $\xi=\{\xi_{1},...,\xi_{M}\}$, where $M$ is the number of layers of AnyTracker network. At the beginning, the zero-initialized adapter does not affect the base tracker’s output. As fine-tuning progresses, the adapter injects dynamics adaptability into the base model through layer-wise feature fusion as shown in Figure~\ref{fig:pipeline}. Such a training paradigm avoids degrading the already acquired motion tracking performance, ultimately yielding both rich motion expressiveness and strong dynamics adaptability.

We use the same rewards in Table~\ref{table:rewards} to finetune the adapter, but within an environment with dynamics variance. In each iteration, we first update the data replay buffer, and then the adapter and the history encoder are alternately updated.

\section{EXPERIMENTS}

In this section, we provide extensive experimental results in both the MuJoCo~\cite{todorov2012mujoco} simulator and the real-world deployment. We choose the 29-DoF Unitree G1 humanoid robot for all our experiments. Here, we aim at addressing the following three questions:

\begin{itemize}
    \item \textbf{Q1}: Can AnyTracker improve the performance of general motion tracking compared to existing methods?
    \item \textbf{Q2}: Can AnyAdapter outperform other baseline methods in online dynamics adaptation capability against disturbances from different sources?
    \item \textbf{Q3}: How does Any2Track perform in various real-world scenarios?
\end{itemize}

\begin{figure*}
    \centering
    \vspace{0.1cm}
    \includegraphics[width=1.0\linewidth]{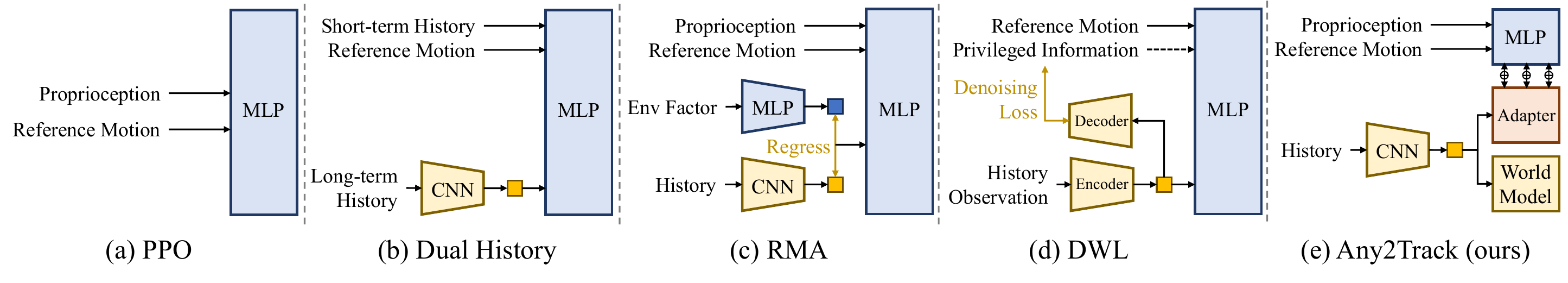}
    \caption{\textbf{Overview of methods evaluated in online dynamics adaptation experiments.} All algorithms are trained with asymmetric PPO, and the critics are omitted in the figure.}
    \label{fig:comparison}
\vspace{-0.5cm}
\end{figure*}

\subsection{Results of Tracking Any Motions}

To address \textbf{Q1} (\textit{Can AnyTracker improve the performance of general motion tracking compared to existing methods?}), we compare AnyTracker with baseline methods and evaluate each of our major designs in this part.

\subsubsection{Experiment Setting}
We use our curated motion datasets, as mentioned in Section~\ref{sec:anytracker}, to train all the methods in simulation. We evaluate on the AMASS test set and LAFAN1 following~\cite{chen2025gmt}. In this experiment, we do not introduce any dynamics variance, only compare the quality of motion tracking.

\subsubsection{Experiment Metrics}
\label{sec:metrics}
We use the following metrics:
\begin{itemize}
    \item \textbf{Success Rate (SR, \%)} records the percentage of successful tracking trials. For motion tracking of a certain trajectory, the tracking is defined as unsuccessful when the averaged joint position error or the root height error is greater than 0.2m.
    \item \textbf{Mean Per Joint Position Error (MPJPE, mm)} is the average position error of all links.
    \item \textbf{Mean Per Joint Velocity Error (MPJVE, mm/frame)} is the average velocity error of all links.
\end{itemize}

\subsubsection{Baselines}
Since different methods utilize different simulators, motion datasets, and even different robots, directly comparing each method is unfair. Thus, we choose existing general motion trackers with open-source training code as our baseline methods and reproduce them in MuJoCo using our curated motion datasets. In this experiment, we re-implement OmniH2O and ExBody2.

We also ablate our major designs in AnyTracker, including canonicalized action spaces (\textbf{Ours w/o CAS}) and specialist-to-generalist distillation (\textbf{Ours w/o distillation}), to validate their effectiveness.

\subsubsection{Experiment Results}
The results are shown in Table~\ref{tab:track_comparison}. AnyTracker outperforms baseline methods in motion tracking quality because of a series of careful designs. The effectiveness of canonicalized action spaces and specialist-to-generalist distillation is also validated. These two designs help alleviate the optimization difficulty brought by complex action spaces and improve the final quantitative results.

\subsection{Results of Adapting to Any Disturbances}

To address \textbf{Q2} (\textit{Can AnyAdapter outperform other baseline methods in online dynamics adaptation capability against disturbances from different sources?}), we compare AnyAdapter with baseline methods and evaluate each of our major designs in this part.

\subsubsection{Experiment Setting}
In this experiment, we introduce dynamics variance in the training process of all methods. Training a general motion tracker across LAFAN1 and AMASS under environment disturbances is very challenging, and the specialist-to-generalist distillation process is quite cumbersome. Therefore, all experiments are conducted solely on the LAFAN1 dataset and remove the distillation process. We compare them in the following environments to evaluate their motion tracking quality under disturbances:
\begin{itemize}
    \item \textbf{Terrains}: We use Perlin noise~\cite{perlin1985image} to generate smooth and natural terrains, where we evaluate the terrain adaptability of all methods.
    \item \textbf{External Forces}: We randomly apply a force of random magnitude and direction to the robot's torso during motion tracking.
    \item \textbf{Physical Property Changes}: We randomly change the torso mass, torso CoM position, and joint friction of the robot.
\end{itemize}

Besides, we also compare them in an environment without any disturbances to evaluate their basic motion tracking capability. 

\begin{table}[]
\vspace{0.2cm}
\caption{\textbf{Comparison with baseline motion tracking methods.} We use our curated motion datasets to evaluate existing general motion tracking methods and our major designs. \textbf{Bold} numbers indicate the best performance.}
\centering
\setlength{\tabcolsep}{10pt}
\begin{tabular}{lccc}
\hline
Method  & SR↑ & MPJPE↓ & MPJVE↓ \\ \hline
OmniH2O &  75.64  &  36.12     &   12.24   \\
Exbody2 &  79.68  &   34.70    &   13.69    \\ \hline
Ours w/o CAS & 84.92 & 30.15 & 9.87 \\
Ours w/o distillation & 83.32 & 31.29 & 9.61 \\
\textbf{Ours}   & \textbf{89.23} & \textbf{27.96} & \textbf{6.43} \\ \hline
\end{tabular}
\label{tab:track_comparison}
\vspace{-0.5cm}
\end{table}

\begin{table*}
\centering
\vspace{0.2cm}
\caption{\textbf{Simulation performance of compared methods on online dynamics adaptation.} \textbf{Bold} numbers indicate the best performance.}
\label{tab:adaptation}
\setlength{\tabcolsep}{4pt}
\begin{tabular}{lccccccccccccccc}
\hline
\multirow{2}{*}{Method} & \multicolumn{3}{c}{\textbf{w/o Disturbance}}   & \multicolumn{1}{l}{\multirow{2}{*}{}} & \multicolumn{3}{c}{\textbf{Terrains}}    & \multicolumn{1}{l}{\multirow{2}{*}{}} & \multicolumn{3}{c}{\textbf{External Forces}}      & \multicolumn{1}{l}{\multirow{2}{*}{}} & \multicolumn{3}{c}{\textbf{Physical Property Changes}} \\ \cline{2-4} \cline{6-8} \cline{10-12} \cline{14-16} 
                        & SR↑           & MPJPE↓         & MPJVE↓        & \multicolumn{1}{l}{}                  & SR↑           & MPJPE↓         & MPJVE↓        & \multicolumn{1}{l}{}                  & SR↑           & MPJPE↓         & MPJVE↓         & \multicolumn{1}{l}{}                  & SR↑              & MPJPE↓           & MPJVE↓          \\ \hline
PPO                     & 87.4          & 28.30          & 7.65          &                                       & 73.0          & 31.01          & 8.99          &                                       & 50.0          & 37.60          & 12.06          &                                       & 73.6             & 34.14            & 8.64            \\
RMA (Stage 1)           & 88.8          & 18.58          & 6.54          &                                       & 75.8          & 23.99          & 8.45          &                                       & 52.2          & 30.22          & 11.18          &                                       & 77.8             & 29.54            & 8.29            \\
RMA (Stage 2)           & 88.4          & 19.68          & 6.76          &                                       & 75.8          & 24.60          & 8.56          &                                       & 51.8          & 30.63          & 11.33          &                                       & 77.8             & 29.52            & 8.35            \\
DWL                     & 87.4          & 26.04          & 7.83          &                                       & 78.6          & 29.27          & 9.31          &                                       & 42.6          & 37.24          & 11.92          &                                       & 70.6             & 35.11            & 9.61            \\
Dual History            & 87.4          & 22.62          & 6.92          &                                       & 79.6          & 26.38          & 8.58          &                                       & 43.6          & 34.29          & 11.62          &                                       & 74.8             & 32.06            & 8.76            \\ \hline
Ours w/o Adapter        & 89.6          & 19.12          & 6.42          &                                       & 82.4          & 22.36          & 7.90          &                                       & 57.6          & \textbf{28.91} & 11.16          &                                       & \textbf{80.6}    & 29.08            & 8.61            \\
Ours w/o World Model    & 88.8          & 21.42          & 6.56          &                                       & 82.6          & 26.32          & 8.21          &                                       & 49.8          & 36.43          & 11.40          &                                       & 70.4             & 32.09            & 8.52            \\
\textbf{Any2Track} (ours)        & \textbf{89.8} & \textbf{16.46} & \textbf{6.04} &                                       & \textbf{83.2} & \textbf{20.68} & \textbf{7.82} &                                       & \textbf{59.0} & 28.97          & \textbf{10.81} &                                       & \textbf{80.6}    & \textbf{27.75}   & \textbf{8.15}   \\ \hline
\end{tabular}
\vspace{-0.5cm}
\end{table*}

\subsubsection{Experiment Metrics}
We adopt the same metrics in Section~\ref{sec:metrics}.

\subsubsection{Baselines}

We choose the following methods as baselines:
\begin{itemize}
    \item \textbf{Vanilla PPO}: We directly train the policy with the vanilla PPO algorithm as we described in AnyTracker, with dynamics variance introduced during training. The policy cannot observe the history or the disturbance.
    \item \textbf{Dual History}~\cite{li2025reinforcement}: The policy can receive both long-term and short-term history as observations. The long-term history is encoded by a convolutional network and concatenated with the short-term history. 
    \item \textbf{RMA}~\cite{kumar2021rma}: It utilizes a two-stage framework to adapt for different environments. In stage 1, the privileged environment factor is encoded by an MLP to form the environment embedding, and the embedding is fed into the policy network and trained with PPO. In stage 2, the policy network is fixed, and a history encoder is trained to regress the environment embedding. We report the performance of RMA in both stages.
    \item \textbf{DWL}~\cite{gu2024advancing}: An encoder-decoder structure is applied to predict privileged information, including privileged robot states and environment dynamics configurations, from the observation history. The decoder is discarded during inference, and the history embedding from the encoder is fed into the policy network.
\end{itemize}

A brief overview of these methods is depicted in Figure~\ref{fig:comparison}. Moreover, we conduct ablation experiments to evaluate the effectiveness of policy adaptation and world model learning in the AnyAdapter. \textbf{Ours w/o Adapter} train the base policy from scratch with history encoder and world model, without the finetuning stage and adapter architecture. \textbf{Ours w/o World Model} train the adapter and history encoder simultaneously via the PPO loss without introducing the world model to provide informative dynamics embeddings. 

\subsubsection{Experiment Results}
The results are exhibited in Table~\ref{tab:adaptation}. We observe that Any2Track surpasses all baseline methods under all disturbances. When there is no disturbance, our method achieves the highest success rate and lowest tracking error.
When dynamic disturbances are introduced into the test environment, our method shows less performance drop on motion tracking compared with the baseline methods, which indicates our better adaptability. Notably, our method even outperforms the RMA (Stage 1), which takes privileged environment factors as input. This is mainly because Any2Track provides better dynamics features and an adaptable learning strategy.

As for ablation studies, we observe that the world model plays an important role in our method for providing informative dynamics embeddings. The tracking performance without the world model is even worse than vanilla PPO when external push or physical property change is applied. Our method without the adapter also suffers a clear performance drop, which validates the effectiveness of decoupling the learning of basic action execution and adaptability.

\subsection{Real-World Evaluations}

\begin{figure}
    \centering
    \vspace{0.1cm}
    \includegraphics[width=1.0\linewidth]{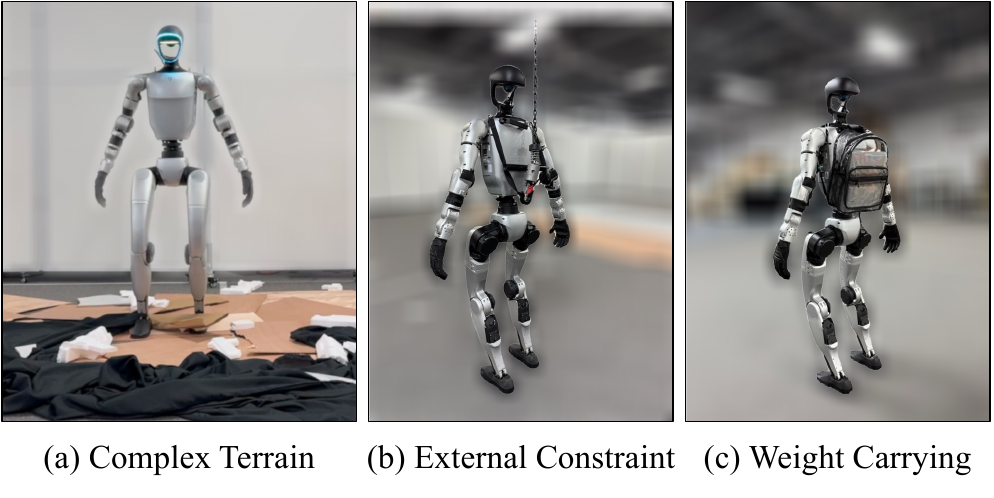}
    \caption{\textbf{Different environment disturbance settings in the real-world experiment.}}
    \label{fig:realworld}
\vspace{-0.5cm}
\end{figure}

\begin{table}
\centering
\vspace{0.2cm}
\caption{\textbf{Real-world performance of compared methods.} Methods are evaluated on identical trajectories and settings.}
\label{tab:realworld}
\setlength{\tabcolsep}{3pt}
\begin{tabular}{lccccc}
\hline
\multirow{2}{*}{Method} & \multicolumn{2}{c}{PPO} & \textbf{} & \multicolumn{2}{c}{Any2Track}   \\ \cline{2-3} \cline{5-6} 
                        & MPJPE↓     & MPJVE↓     &           & MPJPE↓         & MPJVE↓         \\ \hline
w/o Disturbance         & 29.38      & 14.26      &           & 17.38 \color[rgb]{0.0,0.5,0.0}(-12.00) & 11.56  \color[rgb]{0.0,0.5,0.0}(-2.70) \\
Complex Terrain          & 37.21      & 15.75      &           & 18.34 \color[rgb]{0.0,0.5,0.0}(-18.87)  & 11.87 \color[rgb]{0.0,0.5,0.0}(-3.88)  \\
External Constraint     & 39.84      & 16.01      &           & 19.17 \color[rgb]{0.0,0.5,0.0}(-20.67)  & 13.04 \color[rgb]{0.0,0.5,0.0}(-2.97)  \\
Weight Carrying         & 37.52      & 16.91      &           & 23.24 \color[rgb]{0.0,0.5,0.0}(-14.28)  & 12.69 \color[rgb]{0.0,0.5,0.0}(-4.22)  \\ \hline
\end{tabular}
\vspace{-0.5cm}
\end{table}

To address \textbf{Q3} (\textit{How does Any2Track perform in various real-world scenarios?}), we deploy our policy on the real Unitree G1 robot and report the results in this part.

\subsubsection{Experiment Setting}
We quantify the performance of Any2Track on real-world robots. As we show in Figure~\ref{fig:realworld}, we conduct real-world experiments under various disturbance conditions, including (a) complex terrain: we use wooden boards, cardboard, foam, and fabric as terrains; (b) external constraint: the robot’s back was connected to a hoist via a fixed-length rope as external forces; and (c) weight carrying: we add a 5 kg payload on the robot's back as physical property changes in robot mass. All results are averaged over 5 different motion trajectories.

\subsubsection{Experiment Metrics}

We report MPJPE and MPJVE (mentioned in Section~\ref{sec:metrics}) in this experiment. We collect joint position/velocity information from the motor sensors and calculate the metrics using forward kinematics.

\subsubsection{Baselines}

We choose vanilla PPO with domain randomization as our baseline method. 

\subsubsection{Experiment Results}

Table~\ref{tab:realworld} depicts the real-world deployment results. The performance of Any2Track beats PPO with domain randomization in all test environments. 
We found that the advantage of Any2Track increases as the environment disturbances are introduced.

\section{CONCLUSIONS}
We propose Any2Track, a novel two-stage motion tracking framework for humanoid robots. 
Any2Track can track diverse, highly dynamic, and contact-rich motions under multiple real-world disturbances, including terrains, external forces, and physical property changes. Our work paves the way for a foundational humanoid motion tracker in open-world environments with general practical use. For future research, Any2Track can serve as a robust base motion tracking model for different downstream tasks, including whole-body tele-operation, humanoid skill learning, humanoid VLA model, and so on.

\bibliographystyle{IEEEtran}
\bibliography{IEEEabrv}

\end{document}